\documentclass[letterpaper]{article} 
\usepackage{aaai2026}  
\usepackage{times}  
\usepackage{helvet}  
\usepackage{courier}  
\usepackage[hyphens]{url}  
\usepackage{graphicx} 
\usepackage{natbib}  
\usepackage{caption} 
\usepackage{algorithm}
\usepackage{algorithmic}

\usepackage{multirow}
\usepackage{booktabs} 
\usepackage{tcolorbox}
\usepackage{newfloat}
\usepackage{listings}
\DeclareCaptionStyle{ruled}{labelfont=normalfont,labelsep=colon,strut=off} 
\floatstyle{ruled}
\newfloat{listing}{tb}{lst}{}
\floatname{listing}{Listing}
\title{Learning from Reasoning Failures via Synthetic Data Generation}
\author{
    Gabriela Ben Melech Stan\textsuperscript{\rm 1},
    Estelle Aflalo\textsuperscript{\rm 1},
    Avinash Madasu\textsuperscript{\rm 1},
    Vasudev Lal\textsuperscript{\rm 2},
    Phillip Howard\textsuperscript{\rm 3}
}
\affiliations{
    \textsuperscript{\rm 1}Intel Labs\\
    \textsuperscript{\rm 2}Oracle\\
    \textsuperscript{\rm 3}Thoughtworks\\

    \{gabriela.ben.melech.stan, estelle.aflalo, avinash.madasu\}@intel.com, vasudev.lal@oracle.com, phillip.howard@thoughtworks.com
}

\begin{document}
\maketitle
\begin{abstract}

Training models on synthetic data has emerged as an increasingly important strategy for improving the performance of generative AI. This approach is particularly helpful for large multimodal models (LMMs) due to the relative scarcity of high-quality paired image-text data compared to language-only data. While a variety of methods have been proposed for generating large multimodal datasets, they do not tailor the synthetic data to address specific deficiencies in the reasoning abilities of LMMs which will be trained with the generated dataset. In contrast, humans often learn in a more efficient manner by seeking out examples related to the types of reasoning where they have failed previously. Inspired by this observation, we propose a new approach for synthetic data generation which is grounded in the analysis of an existing LMM's reasoning failures. Our methodology leverages frontier models to automatically analyze errors produced by a weaker LMM and propose new examples which can be used to correct the reasoning failure via additional training, which are then further filtered to ensure high quality. We generate a large multimodal instruction tuning dataset containing over 553k examples using our approach and conduct extensive experiments demonstrating its utility for improving the performance of LMMs on multiple downstream tasks. Our results show that models trained on our synthetic data can even exceed the performance of LMMs trained on an equivalent amount of additional real data, demonstrating the high value of generating synthetic data targeted to specific reasoning failure modes in LMMs.

\end{abstract}  
\section{Introduction}
\label{sec:intro}

Recent advancements in large language models (LLMs) have significantly advanced the state-of-art across a broad range of problem domains, which can be attributed in large part to scaling the size of models and datasets used for training. However, a limiting factor to further scaling of training datasets is the availability of new high-quality data. With recent studies suggesting that we are rapidly approaching saturation of existing public data sources \citep{villalobos2022will,xue2023repeat}, synthetic data generation has become an increasingly popular approach for augmenting training datasets. The generation of high-quality synthetic data has become more feasible recently as the capabilities of LLMs have increased, thereby enabling their use for the production of synthetic data at scale \citep{ding2024data, yu2023large, lee2024llm2llm, howard2022neurocounterfactuals, howard2023neurocomparatives, su2024sk}.

Synthetic data has been especially valuable for training large multimodal models (LMMs), which combine an LLM with a vision encoder to enable text generation conditioned on multimodal inputs. Unlike LLMs, real data for training LMMs is scarce due to the lack of naturally-occurring images paired with high-quality text. Consequently, popular LMMs such as LLaVA \citep{liu2023visual} use partially synthetic data for multimodal instruction tuning, which is often collected by generating synthetic text for real images. 

\begin{figure*}
    \centering
    \includegraphics[width=\linewidth]{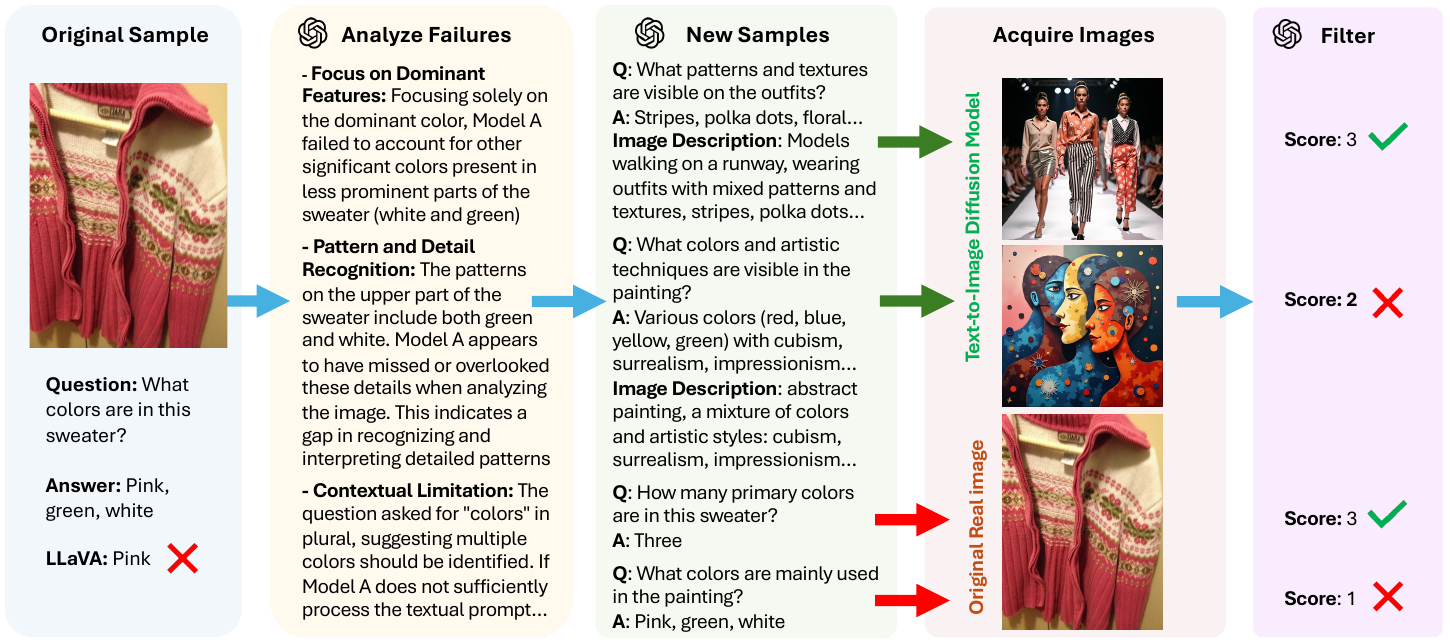}
    \caption{Illustration of our approach. Given a sample from an existing dataset which LLaVA answers incorrectly, we prompt a frontier model to analyze LLaVA's reasoning failures and propose new synthetic samples which require similar types of reasoning. }
    \label{fig:intro-figure}
\end{figure*}

Existing approaches to generating synthetic data suitable for training LMMs suffer from two primary limitations. First, most rely on the acquisition of real images, which are then paired with synthetic text produced by another LMM. This limits their application to domains where images are readily available and therefore makes such approaches ill-suited for low-resource settings where image data is scarce. Second, prior approaches utilize broad generation strategies which arbitrarily produce synthetic examples without consideration of the type of data which would be most useful for improving LMMs. This results in an inefficient approach for both generating data and training LMMs, as many of the synthetic examples may provide little or no incremental value when added to training datasets.

In contrast to the dominant paradigm for generating synthetic training data, humans often learn more efficiently by focusing their attention on examples which are most related to their past reasoning failures. To acquire expert knowledge, \citet{ericsson1993role} argue that humans should ``ideally be given explicit instructions about the best method and be supervised by a teacher to allow individualized diagnosis of errors, informative feedback, and remedial part training.'' Since humans acquire new knowledge by learning from their failures 
\citep{darabi2018learning,brown1989situated,morgan1904introduction}, 
seeking out additional examples related to past reasoning errors for further practice and evaluation helps the learner achieve expertise in a subject. Problems which require types of reasoning that have already been mastered are often ignored under this paradigm, as they provide little incremental value in the learning process. 

Motivated by such observations of human learning behavior, we propose a new approach for generating synthetic multimodal data which is grounded in an existing LMM's reasoning failures. First, we identify the failure modes of an existing LMM by evaluating it on the training split of a benchmark dataset and then asking a strong frontier model to hypothesize possible reasons for its incorrect predictions. The frontier model then proposes new question \& answer pairs related to the identified reasoning failure modes as well as descriptions of images to accompany them, which can be either used to link the new examples to existing images, or as prompts to generate synthetic images. We further filter the resulting synthetic data using an LMM-as-a-judge methodology to ensure high quality.

Using our approach, we generate a large multimodal instruction tuning dataset containing over 553k examples derived from the reasoning failures of LLaVA-1.5-7B, a popular LMM. We conduct extensive training experiments showing that our synthetic data improves LLaVA's performance on multiple downstream tasks, even outperforming additional training on alternative real datasets. This is notable, as prior studies have shown that training on synthetic data often requires significantly more examples to achieve equivalent results as training on real data \citep{he2022synthetic}. Training on our dataset also produces greater performance gains than previously-proposed synthetic datasets, highlighting the utility of grounding synthetic data generation in the analysis of a model's reasoning failures. 

\noindent To summarize, our contributions are as follows:
\begin{enumerate}
    \item We propose a new approach for generating fully synthetic multimodal datasets which is grounded in an analysis of an existing LMM's reasoning failures.
    \item Using our methodology, we produce a large multimodal instruction tuning dataset containing over 553k examples derived from errors produced by LLaVA.
    \item We conduct extensive experiments demonstrating the utility of our dataset for improving LMMs via training.
    \item Through ablations and human analyses, we demonstrate the high quality of data produced by our approach.
\end{enumerate}
\section{Related Work}
\label{sec:related}

\paragraph{Synthetic Datasets for Training LMMs.}
\citet{chen2024allava} proposed the ALLaVA dataset, which consists of 1.3M samples of real images paired with image annotations and question-answer pairs generated by a frontier LMM. While their dataset is similarly suited for multimodal instruction tuning as our dataset, they utilize only real images and do not ground data generation in model failures, whereas our approach can utilize both real and synthetically generated images. \citet{li2024synthesize} generate synthetic question-answer pairs from an LLM for real chart images to address the narrower domain of reasoning-based chart VQA. \citet{yang2025scaling} generate synthetic multimodal data for text-rich images, leveraging code-guided generation in languages such as LaTeX and HTML to automatically generate synthetic images with dense text descriptions. In contrast, our approach can be broadly applied to different problem domains while leveraging text-to-image diffusion models to produce more diverse types of images.

Other approaches to generating synthetic data suitable for training MLLMs differ from ours in the types of reasoning which they target. Img-Diff~\citep{jiao2025img} contains contrasting synthetic examples which differ only in terms of certain objects in specific regions of an image, requiring the MLLM to learn to identify and articulate such differences. Data-Juicer 2.0~\citep{chen2024data} provides multimodal synthetic data generation capabilities utilizing foundation MLLMs and text-to-image diffusion models, but does not address our specific task of generating synthetic data derived from reasoning failures. For a comprehensive data-centric review of MLLMs and how they can assist in the data generation process, see \citet{qin2025synergy}.

\paragraph{Synthetic Data Generation from Model Failures.}
Some prior work has explored the use of model failures in the context of synthetic data generation. \citet{jain2022distilling} proposed an approach to identify directions in a vision model's latent space which correspond to model failures, which can then be integrated into diffusion models to generate synthetic images tailored to correcting the model's mistakes. \citet{chegini2023identifying} use ChatGPT and CLIP to identify text descriptions corresponding to a vision model's failure modes, which are used as prompts for diffusion models to produce synthetic images for data augmentation. In the realm of language-only models, DISCERN \citep{menon2024discern} utilizes two LLMs to iteratively produce natural language descriptions of errors for synthetic data generation. \citet{lee2024llm2llm} use incorrect answers from a student LLM finetuned on specific tasks as input to a teacher LLM which generates new examples to use for training. These prior works differ from ours in that they (1) generate only single-modality data (e.g., text-only or image-only), and (2) focus on classification tasks such as image recognition or text classification. In contrast, our approach generates image-text datasets aimed at training models for open-ended text generation conditioned on multimodal inputs.


\paragraph{Generating Synthetic Data from Frontier Models to Teach New Skills.}
AgentInstruct \citep{mitra2024agentinstruct} is an agentic framework for generating synthetic data from a powerful frontier model (e.g., GPT-4) to teach new skills to a weaker LLM. Similarly, \citet{ziegler2024craft} utilize few-shot examples annotated by humans and retrieved documents with produce synthetic data from LLMs for teaching specialized tasks to models. Prompt-based methods for synthetic data generation from LLMs without seed documents \citep{gupta2023targen,ubani2023zeroshotdataaug} as well as knowledge distillation from a teacher model \citep{kim2024promptkd} have also been proposed. Unlike our work, these prior studies focus on language-only data generation and use seed documents (e.g., raw text, source code) or prompts as a basis for data generation rather than an analysis of model failures.

\section{Dataset Construction}
\label{sec:method}

\subsection{Diagnosing Model Failures}
To generate our tailored synthetic data, we first analyze reasoning failures in a baseline LMM using a more advanced frontier LMM. The frontier model is selected for its superior multimodal reasoning capabilities and high accuracy on diverse vision-language benchmarks. Reasoning failures are identified by evaluating both models on the training sets of vision-language benchmarks, and selecting samples where the baseline LMM produces incorrect responses, while the frontier model succeeds. This process generates a subset of failure cases per benchmark, which we denote as Model Failure Sets (MFS). Each MFS highlights the weaknesses of the baseline LMM, providing a focused challenging dataset.


\subsection{Synthetic Data Generation}
\label{sec:synthetic-data-gen}
Using the resulting MFS, we guide the frontier model through a structured multi-turn process to diagnose the failures of the baseline LMM and generate new training samples to address these failure modes.


The frontier model first analyzes the image, question, ground truth, and incorrect prediction to diagnose reasoning errors. It is then instructed to propose new challenging samples consisting of a detailed image description, question, and deterministic answer designed to address the failures. 
We explore two sourcing strategies for the images: real images and synthetically generated ones. These two methods utilize the prompt described in Figure 5 of the Appendix. For Method 1, the final image-generation step is skipped.

\paragraph{Method 1: Question-Answer Generation for Existing Images.} We leverage the original images from failed samples and prompt the frontier model to generate 10 new question-answer pairs per sample using the reasoning pipeline described above. This approach is particularly effective for benchmarks like InfoVQA and ScienceQA, where text-to-image models often fail to capture fine-grained text and generate accurate spatial details.
\paragraph{Method 2: Fully Synthetic Question, Answer, and Image Generation.} In this approach, the frontier model generates both a question-answer pair and a detailed image description, which is used to prompt a text-to-image diffusion model to generate synthetic images. For each failed sample, we create 100 fully synthetic samples, consisting of 10 (image prompt, question, answer) triplets, with each image prompt used to produce 10 images using different classifier-free guidance scales to enhance image diversity.







\begin{figure}[h] 
    \centering
    \includegraphics[width=0.8\columnwidth]{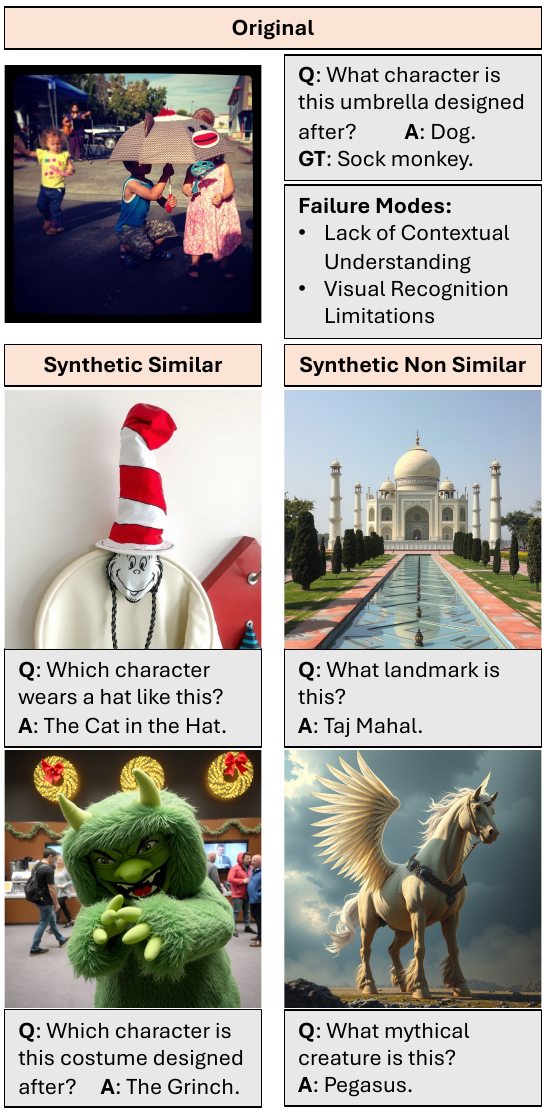}
    \caption{Comparison of fully synthetic similar and non-similar samples. Similar samples maintain a children's characters-based theme like the original sample, while non-similar samples address the failure modes by introducing diverse contexts.
    }
    \label{fig:similar}
\end{figure}

\paragraph{Encouraging Diversity.} To further enhance the diversity of our data, we use a variation of our prompt which instructs the frontier LMM to ``provide examples that challenge Model A's weaknesses using scenarios from entirely different domains.'' This additional instruction is appended to Step 4 (Figure 5) to encourage the generation of samples in different domains and enhance generalization.
Figure \ref{fig:similar} compares fully synthetic data with similar and non-similar domains. We also varied constraints on the question format (e.g., multiple-choice, true/false) and answer instructions (e.g., requiring "Unanswerable" or brief responses, see the Shiba Inu example in Figure 10).


\paragraph{Filtering}
\label{sec:filtering}
We use the same frontier LMM to filter synthetic samples by rating image-question-answer triplets from 1 (incorrect) to 3 (fully correct), as shown in the prompt in Figure 6. Only samples rated 3 are retained to ensure quality.


\begin{figure}[h] 
    \centering
     \includegraphics[width=0.8\columnwidth]   {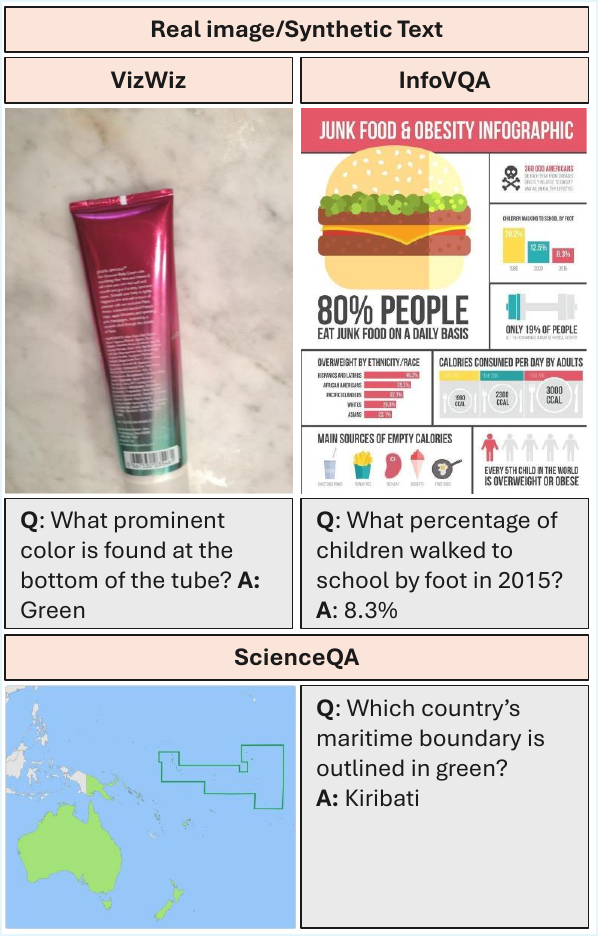}
    \caption{Examples of generated synthetic question-answer pairs for real images from VizWiz, InfoVQA, and ScienceQA.}
    \label{fig:examples_dataset_realim}
\end{figure}

\subsection{Details of Generated Dataset}
\label{sec:dataset-details}

Our synthetic dataset consists of 553,992 samples incorporating both real and generated images derived from the MFS of LLaVA-1.5-7B on four benchmark training sets: VizWiz \citep{gurari2018vizwiz}, InfoVQA \citep{mathew2022infographicvqa}, ScienceQA \citep{lu2022learn}, and OK-VQA \citep{marino2019ok}. These benchmarks were chosen to cover a wide range of visual reasoning challenges. 
VizWiz features real-world images from visually impaired users, demanding detailed scene understanding. OK-VQA requires external knowledge for visual question answering. InfoVQA focuses on text-rich images, testing reading comprehension. ScienceQA involves multimodal scientific questions, assessing spatial and logical reasoning skills.
We use GPT-4o \cite{openai2024gpt4ocard} as the frontier LMM  for analyzing the reasoning failures of LLaVA-1.5-7B due to its strong multimodal reasoning capabilities.
The text-to-image model FLUX.1-schnell \citep{flux2024} was used to generate the images with a resolution of 1024×1024 pixels and a classifier guidance scale ranging from 3 to 13.

Table 6 (Appendix) provides an overview of each benchmark, including the size of the training split, the MFS size, and the number of synthetic samples after filtering.
For InfoVQA and ScienceQA, we applied Method 1 from Section~\ref{sec:synthetic-data-gen}. For OK-VQA, we employed Method 2. For the VizWiz, we integrated both methods. 
The final dataset comprises 42\% real-image-based and 58\% fully synthetic samples, each with a single-turn question-answer pair.

Our filtering process effectively removes low-quality samples. VizWiz had the highest rejection rate (81\% for synthetic images and 34\% for real images), highlighting the challenge of generating quality synthetic visuals. OK-VQA’s lower rate (29\%) suggests simpler and less ambiguous visual content. ScienceQA showed a similar 29\% rejection rate for real images, while InfoVQA had only 5\% removed.
Figure~\ref{fig:examples_dataset_realim} showcases diverse question types and the quality of generated samples from VizWiz and InfoVQA. Appendix D includes additional examples, including samples with errors or ambiguities. Finally, Figure 8 shows OK-VQA examples with original image-question-answer pairs, the GPT-4o-identified failure mode, and newly generated samples targeting the same failure.
\section{Experiments}
\label{sec:experiments}

\subsection{Details of Experimental Setting}

We perform vision-language instruction tuning with LLaVA-v1.5-7B, building upon the LLaVA-1.5-mix-665K instruction tuning dataset \cite{liu2024improvedbaselinesvisualinstruction}. This dataset contains 665K structured user-GPT conversations mostly focused on visual prompts. Because our approach explicitly targets image-grounded reasoning, we only selected the 624K samples that include images from the original LLaVA-instruct data. We then augmented this visually oriented subset with our generated synthetic MFS (Section~\ref{sec:dataset-details}), varying the number of synthetic samples used according to the requirements of each experiment. The resulting training set combines real image-text conversations and generated synthetic samples that address specific reasoning failures.
For fine-tuning, we used Vicuna-v1.5-7B \cite{zheng2023judgingllmasajudgemtbenchchatbot} weights as the LLM backbone and leveraged the pretrained multimodal projector from LLaVA-1.5-7b.
We followed the original LLaVA training procedure, ensuring a fair comparison to existing methods (details in Appendix A.5).
We refer to models trained with a mixture of the original LLaVA-Instruct dataset and our synthetic data as $\textrm{LLaVA}_{\textrm{syn}}$. As a baseline, we report the performance of LLaVA trained under the same setting but without any additional synthetic data added to the training dataset (i.e., $N_{syn} = 0$). Additionally, for each dataset from which reasoning failures were derived for synthetic data generation, we report the performance of a LLaVA model trained on an equivalent amount of real data sourced from the corresponding training dataset (denoted as $\textrm{LLaVA}_{\textrm{real}}$). This provides a measure of the efficiency of our synthetic data relative to training on real in-domain data. 

\subsection{Synthetic Data Augmentation Results}

Table~\ref{tab:training-data-augmentation-in-domain} provides in-domain evaluation results utilizing synthetic data derived from InfoVQA, ScienceQA, and OK-VQA. 
Notably, augmenting the LLaVA-Instruct dataset with our synthetic data achieves performance comparable to or better than using an equivalent amount of real domain-specific data in most cases. 
This result is particularly significant given that the synthetic samples were generated using only a small subset of the original training data: specifically, only those examples where LLaVA scored 0.0 while GPT scored 1.0. 
For instance, in OK-VQA, the original training set consists of 9,009 samples, but we utilized only 607 training samples where LLaVA failed in order to generate 9009 synthetic samples, resulting in a performance boost of 13\% on the OK-VQA test set. Similarly, our approach utilized only 28\% of the ScienceQA training dataset to generate full synthetic replacements, yet still resulted in better performance than training directly on the real dataset. 
We also observe that performance improves as the amount of synthetic data used for data augmentation increases.

In practice, it may be desirable to combine synthetically generated data which was derived from reasoning failures across different datasets. We therefore provide results for two different sized mixtures of our synthetic data derived from InfoVQA, ScienceQA, and OK-VQA reasoning failures in Table~\ref{tab:training-data-augmentation-mix-all-other-datasets}. We also provide results for LLaVA models trained on four alternative synthetically generated datasets: ALLaVA \cite{chen2024allava}, CoSyn-400k \cite{yang2025scaling}, SimVQA \cite{cascantebonilla2022simvqaexploringsimulatedenvironments}, and Img-Diff \cite{jiao2024imgdiffcontrastivedatasynthesis}\footnote{ALLaVA consists of real images with synthetic text, whereas Cosyn-400k and SimVQA contain fully synthetic examples. Our synthetic data utilized here contains both real and synthetic images.}.
From Table~\ref{tab:training-data-augmentation-mix-all-other-datasets}, we observe that our synthetic data outperforms these baselines datasets, demonstrating the value of grounding synthetic data generation in an analysis of a model's failures. 
Finally, we evaluated trained models on OOD benchmarks (Table 8 of Appendix) and found that models trained on our dataset do not exhibit degradation in OOD performance, demonstrating the ability of our approach to correct model reasoning failures without sacrificing general reasoning capabilities. Importantly, our dataset strikes a middle ground in the cost–quality–diversity tradeoff and can be further complemented with lower-cost image generation pipelines such as CoSyn’s to achieve scalable and semantically rich supervision. A detailed comparison of compute and generation costs across our method and baseline datasets is provided in Appendix B.3.

\begin{table}[]
    \centering
    \fontsize{9}{11}\selectfont
    {
    \begin{tabular}{llcccc}
    \toprule
    Dataset & Model & $N$ & $N_{syn}$ & EM Score \\
    \midrule
    \multirow{5}{*}{I-VQA} & LLaVA & 624,610 & 0 & 26.7 \\
    & $\textrm{LLaVA}_{\textrm{real}}$ & 634,684 & 0 & 31.6 \\
    & $\textrm{LLaVA}_{\textrm{syn}}$ & 634,684 & 10,074 & 30.8\\
    & $\textrm{LLaVA}_{\textrm{syn}}$ & 687,071 & 62,461 & 33.0 \\
    & $\textrm{LLaVA}_{\textrm{syn}}$ & 710,610 & 86,000 & \textbf{34.3}\\
    \midrule
    \multirow{4}{*}{SQA} & LLaVA & 624,610 & 0 &  70.7\\
    & $\textrm{LLaVA}_{\textrm{real}}$ & 630,195 & 0 &  70.0\\
    & $\textrm{LLaVA}_{\textrm{syn}}$ & 630,195 & 5,585 & 71.8\\
    & $\textrm{LLaVA}_{\textrm{syn}}$ & 646,594 & 21,984 &  \textbf{73.0}\\
    \midrule
    \multirow{4}{*}{OK-VQA} & LLaVA & 624,610 & 0 &  57.0\\
    & $\textrm{LLaVA}_{\textrm{real}}$ & 633,619 & 0 &  54.3\\
    & $\textrm{LLaVA}_{\textrm{syn}}$ & 633,619 & 9,009 & 61.3\\
    & $\textrm{LLaVA}_{\textrm{syn}}$ & 687,071 & 62,461 & 61.3\\
    & $\textrm{LLaVA}_{\textrm{syn}}$ & 749,532 & 124,922 & \textbf{61.5}\\
    \bottomrule
    \end{tabular}
    }
    \caption{In-domain evaluation of baseline LLaVA, LLaVA trained using synthetically augmented data ($\textrm{LLaVA}_{\textrm{syn}}$), and augmented with real in-domain data ($\textrm{LLaVA}_{\textrm{real}}$). I-VQA denotes InfoVQA and SQA denotes ScienceQA.} 
    \label{tab:training-data-augmentation-in-domain}
\end{table}

\begin{table}[t]
    \setlength{\tabcolsep}{1.5pt}
    \centering
    \setlength{\tabcolsep}{1mm}
    \fontsize{9}{11}\selectfont
    {
    \begin{tabular}{ccclccccccccc}
    \toprule
    Base LLM & $N$ & $N_{syn}$ & Aug Data & I-VQA & OK-VQA & SQA \\
    \midrule
    \multirow{1}{*}{Vicuna-7B} & 625k & 0 & N/A (bs) & 26.7  & 57.0 & 70.7 \\
    \midrule
    \multirow{2}{*}{Vicuna-7B} & \multirow{2}{*}{643k} & \multirow{2}{*}{18k} & Img-Diff & 27.3 & 56 & 71.3 \\
   & & & Ours & \textbf{31.5} & \textbf{61.2} & \textbf{72} \\
    \midrule
    \multirow{2}{*}{Vicuna-7B} & \multirow{2}{*}{687k} & \multirow{2}{*}{62k} & SimVQA & 26.6 & 54.4 & 71.1 \\ 
   & & & Ours & \textbf{33.1} & \textbf{60.8} & \textbf{73.1} \\
    \midrule
   \multirow{3}{*}{Vicuna-7B} & \multirow{3}{*}{750k} & \multirow{3}{*}{125k} & ALLaVA & 28.8 & 49.4 & 66.5 \\
  &  & & CoSyn & 29.6 & 57.7 & 70.9 \\
   & & & Ours & \textbf{33.2} & \textbf{61.1} & \textbf{73.0} \\
    \midrule
    \midrule
        \multirow{2}{*}{Gemma-2B} & 625k  & 0  & \multirow{2}{*}{Ours} & 21.8  & 51.7 & 62.3 \\
     &  749k & 125k &  & \textbf{29.8} & \textbf{54.8} & 
     \textbf{65.2} \\
    \midrule
   \multirow{2}{*}{Qwen2-7B} & 625k & 0 & \multirow{2}{*}{Ours} & 26.7 & 59.2 & 77.9 \\
         &  750k & 125k &  &\textbf{27} & \textbf{60.6} & \textbf{80.7} \\
    \bottomrule
    \end{tabular}
    }
    \caption{\textbf{Training data augmentation results.} $N$ denotes the total number of training examples, $N_{syn}$ denotes the number of synthetic examples in the training dataset generated using our approach. I-VQA denotes InfoVQA and SQA represents ScienceQA. The first section of the table compared the same base LLM (Vicuna) trained on various datasets with our dataset, while the second section compares different LLM backbones trained on our dataset. 
    } 
    \label{tab:training-data-augmentation-mix-all-other-datasets}
\end{table}


\subsection{Generalization to Other Models and Tasks}

While our synthetic data is collected by targeting the reasoning failures of a specific LMM, we also evaluated whether it could generalize to other models and tasks, thereby further increasing its value beyond correcting targeted reasoning failures in a single model. 
To evaluate this generalization ability of our synthetic dataset, we trained models using the same datasets as before but with different backbone LLMs. In the following experiments, we used Gemma-2B \cite{gemmateam2024gemmaopenmodelsbased} and Qwen2-7B \cite{yang2024qwen2technicalreport} as base LLMs.
We adopted the LLaVA two-phase training procedure: pretraining on the LLaVA 558k dataset \cite{liu2024improvedbaselinesvisualinstruction} followed by instruction fine-tuning. Table \ref{tab:training-data-augmentation-mix-all-other-datasets} indicates that our synthetic data in its maximum augmentation setting, $N_{syn} = 124,922$, outperform the baseline for both LLaVA-Gemma-2B and LLaVA-Qwen-7B across nearly all benchmarks.
Notably, despite being generated based on LLaVA-Vicuna-7B failure modes, our synthetic data enhances the performance of other models whether they are of the same size (Qwen2-7B) or smaller (Gemma-2B). 

For some models (e.g., Qwen2.5-VL-7B-Instruct), the original finetuning dataset is not publicly available. Nevertheless, we show that our dataset can be sufficient to improve these models through continued finetuning. Table~\ref{tab:cft-qwen} provides the results of finetuning Qwen2.5-VL-7B on our 124K synthetic-only dataset derived from LLaVA reasoning failures. We observe that finetuning Qwen2.5-VL-7B on our 124K dataset improves performance significantly on OK-VQA and ScienceQA. InfoVQA performance remains unchanged, likely due to Qwen’s high baseline and prior exposure to similar data. Generating synthetic data which targets Qwen-specific failure cases may yield further gains; nevertheless, these results point to the value of our dataset beyond correcting reasoning failures in only the targeted LMM. 

Finally, we investigated whether our targeted synthetic dataset could improve performance on other tasks besides those used for deriving the reasoning failures. Specifically, we evaluated our $\textrm{LLaVA}_{\textrm{syn}}$ model trained on the 124K synthetic dataset on multiple hallucination and math benchmarks. The results in Table~\ref{tab:llava-hallucination-math} show that $\textrm{LLaVA}_{\textrm{syn}}$ outperforms the baseline LLavA model across all such benchmarks, demonstrating how our synthetic dataset improves model performance beyond the specific reasoning failures targeted during the dataset generation process. The ability of our synthetic dataset to generalize to improving other LMMs and tasks than the ones used in the generation process further illustrates the significant value of our approach. 

\begin{table}[]
    \centering
    \fontsize{9}{11}\selectfont
    \begin{tabular}{lcc}
    \toprule
    & LLaVA & $\textrm{LLaVA}_{\textrm{syn}}$ (124k) \\
    \midrule
    HalluB$\uparrow$ & 40.6 & \textbf{47.4} \\
    LLaVA-B-COCO$\uparrow$ & 85.3 & \textbf{86.9} \\
    POPE$\uparrow$ & 86.8 & \textbf{87.2}\\
    Math-V(vision)$\uparrow$ & 10.6 & \textbf{11.7} \\
    MathVista$\uparrow$ & 24.9 &  \textbf{26.6}\\
    \bottomrule
    \end{tabular}
    \caption{Our targeted 124k synthetic dataset generalizes to improve LLaVA on hallucination and math benchmarks.}
    \label{tab:llava-hallucination-math}
\end{table}

\subsection{Continued Finetuning Results}

We also conducted experiments wherein we continued finetuning LLaVA on our 124K mixture dataset. This approach allows us to evaluate performance gains attributed solely to our data while significantly reducing compute requirements compared to full finetuning. In this setting, we used the same hyperparameter configurations as our previous finetuning experiments. Table \ref{tab:cft-qwen} shows that continued finetuning improves in-domain performance except on OK-VQA. Notably, continued finetuning on the real OK-VQA training set (9K, 41.1\%) results in worse performance than with an identical-size synthetic dataset (48.2\%), suggesting possible overfitting when training on real data.  

\begin{table}[]
    \centering
    \fontsize{9}{11}\selectfont
    \begin{tabular}{@{}lccc@{}}
      \toprule
      & InfoVQA & OK-VQA & SciQA\\
      \midrule
      LLaVA1.5-7b        & 26.4 & \textbf{57.1} & 70.8\\
      + CFT & \textbf{32.4} & 45.8 & \textbf{73.2}\\
    \midrule
      Qwen2.5-VL-7b-Instruct        & 82.6 & 42.4 & 67.9\\
      + CFT & 82.5 & \textbf{53.0} & \textbf{85.5}\\
      \bottomrule
    \end{tabular}%
    \caption{Continued finetuning results for LLaVA-1.5-7b and Qwen2.5-VL-7B-Instruct with our 124K synthetic dataset}
    \label{tab:cft-qwen}
\end{table}

\subsection{Correcting Specific Types of Reasoning Failures}

\begin{table}[]
    \centering
    \begin{tabular}{c|c}
    \toprule
    Cluster name & Percentage (\%) \\
    \bottomrule
       Feature misinterpretation  & 8 \\
        Blurriness & 12.4 \\
        Weakness in visual analysis & 18.6 \\
        Incomplete context understanding & 24\\
        Other & 11.3 \\
        Text recognition errors & 9.2 \\
        Object recognition failure & 10.8 \\
        Overgeneralization & 5.6 \\
        \bottomrule
    \end{tabular}
    \caption{Table shows the clusters of LLAVA reasoning failures described by GPT-4o.}
    \label{tab:reasoning-failure-count}
\end{table}

Our synthetic data generation approach identifies specific types of LMM failures.
To systematically categorize these failures, we encoded each reasoning explanation using sentence transformers ~\cite{reimers-2019-sentence-bert} and clustered them using k-means. Table ~\ref{tab:reasoning-failure-count} presents the resulting clusters, highlighting prevalent failure modes such as optical character recognition (OCR) and object detection errors. Based on this analysis, we further investigated whether targeted synthetic data can effectively address these specific failure cases and enhance LLAVA’s reasoning capabilities.

\begin{table}[t]
    \centering
    \fontsize{9}{11}\selectfont
    \begin{tabular}{lcc}
    \toprule
     Dataset & LLAVA & $\textrm{LLaVA}_{\textrm{syn}}$ \\
     \midrule 
     CIFAR-10  & \textbf{82.1} & 81.2 \\
     Food-101  & \textbf{13.4} & 13.2 \\
     iNaturalist  & 20.6 & \textbf{52.0} \\
     MNIST & 75.1 & \textbf{80.5} \\
     F-MNIST  & 9.8 & \textbf{10.0} \\
     Oxford-pets  & 39.6 & \textbf{96.4} \\
    \bottomrule
    \end{tabular}
    \caption{Image classification results, with $\textrm{LLaVA}_{\textrm{syn}}$ augmented only using examples for object recognition failures.} 
    \label{tab:reasoning-failures-object-recognition}
\end{table}
Specifically, we augmented LLaVA-Instruct with 10,579 synthetic samples from our $\textrm{VizWiz}_{\textrm{syn}}$-MFS addressing object detection reasoning failures and repeated the second stage of LLaVA finetuning. The model was then evaluated on CIFAR-10~\cite{krizhevsky2009learning}, Food-101~\cite{bossard2014food}, iNaturalist~\cite{van2018inaturalist}, MNIST~\cite{lecun1998gradient}, Fashion-MNIST~\cite{xiao2017fashion}, and Oxford-Pets (Binary)~\cite{parkhi2012cats} by formatting samples as multiple-choice questions.
Table~\ref{tab:reasoning-failures-object-recognition} presents a comparison of $\textrm{LLaVA}$ and $\textrm{LLaVA}_{\textrm{syn}}$. $\textrm{LLaVA}_{\textrm{syn}}$ surpasses $\textrm{LLaVA}$ on four out of six datasets, with significant improvements on iNaturalist, MNIST and Oxford-Pets.
This demonstrates the effectiveness of our synthetic dataset in addressing specific reasoning failures within LLaVA. 
Our findings highlight the effectiveness of leveraging targeted synthetic data to refine model reasoning and suggest that incorporating such data-driven interventions can significantly enhance the robustness and generalization of LMMs.

\section{Analysis}
\label{sec:analysis}

\subsection{Human Evaluation of Dataset Quality}

Three of the authors of this work conducted a human evaluation by assessing three different aspects of our generated samples: (1) the alignment of the question and answer in relation to the image prompt, (2) the alignment between the image prompt and the generated image, and (3) the correctness of the answer given the question and image. The first evaluation reflects the semantic coherence, the second evaluates the fidelity of the image generator’s output, and the third combines both aspects. Scores range from 1 to 3, where 1 indicates an irrelevant alignment, 3 signifies a relevant alignment, and 2 represents a partially relevant or ambiguous alignment. 
We evaluated 200 samples in total, with 101 containing real images and 99 being fully synthetic. The overall correctness score for answers was 2.78, with real‐image‐based samples scoring 2.75 and fully synthetic samples scoring 2.81, indicating that fully synthetic samples achieve a level of fidelity equal to or even slightly exceeding that of real‐image-based samples. For the synthetic samples specifically, we also measured the alignment between the image prompt and the generated image (2.66), and the alignment of the generated question and answer with the image prompt (2.84), indicating the high quality of reasoning in the generated responses. Additionally, to assess the frontier model’s reasoning and ensure it highlights meaningful failures, five annotators rated 209 samples (image, QA, and error analysis) on a 1–3 scale. The average score of 2.81 indicates that the explanations usually identified the true failure cause.


\subsection{Impact of LMM on Generated Data}
\label{sec:llm-comparison}
We evaluate our synthetic‐data pipeline with two LMMs of very different cost profiles: the proprietary GPT-4o API and the fully open, Apache-2.0-licensed Qwen2-VL-7B~\cite{wang2024qwen2vlenhancingvisionlanguagemodels}. 
Both models improve LLaVA-7B after fine-tuning on the resulting data: GPT-4o yields gains of 2.7\% on InfoVQA and 10\% on OK-VQA, while Qwen2-VL-7B delivers 7.0\% and 3.5\% on the same tasks.  
Although GPT-4o’s detailed and precise reasoning may contribute to generating more targeted and effective synthetic samples, which produces larger gains on some benchmarks, our failure-guided generation shows robustness by producing meaningful improvements across both models.  
Because Qwen2-VL-7B runs locally on a single A100 GPU, synthetic data can be obtained with our approach at a lower cost than commercial models such as GPT-4o.  
Figure 7 (Section C.1 in Appendix) illustrates representative examples of the reasoning and sample diversity across the two models.

\section{Conclusion}

We introduced a novel method for generating multimodal synthetic data by analyzing model reasoning failures. This approach enabled us to create a multimodal instruction tuning dataset with over 553k synthetic examples derived from LLaVA's errors. Experimental results show that our synthetic data significantly improves LLaVA's in-domain performance on InfoVQA, ScienceQA, and OK-VQA, even outperforming training on an equivalent amount of real data sampled from most of these datasets. Furthermore, models trained on our synthetic dataset exhibit improvements in OOD evaluations and outperform training on other existing synthetic datasets when the amount of training data augmentation is scaled. 
We also showed that training LLaVA only on examples derived from specific failure modes improves its performance on tasks which require corresponding forms of reasoning. 
Through additional ablations and human evaluations, we demonstrated the effectiveness of different components of our methodology and the high quality of our dataset. We believe our study points to the promise of targeted synthetic data generation strategies which leverage an understanding of a model's reasoning deficiencies to construct better synthetic examples.

\label{sec:conclusion}

{
    \small
    \bibliographystyle{aaai2026}
    \bibliography{aaai2026}
}

\clearpage
\setcounter{page}{1}
\appendix
\section*{Learning from Reasoning Failures via Synthetic Data Generation: Supplementary Material}
\section{Implementation details}

\subsection{Compute Infrastructure} 

To generate our dataset, we queried GPT-4o through the Azure OpenAI API and deployed Qwen2-VL on Nvidia RTX A6000 GPUs.  Using
Intel\textsuperscript{\textregistered}\space Gaudi 2 AI accelerators from the Intel\textsuperscript{\textregistered}\space Tiber\textsuperscript{\texttrademark}\space AI Cloud, we generated 1.024 million images from the VizWiz failed samples and 535k images derived from OK-VQA.

\subsection{Dataset statistics}
\label{app:dataset-statistics}

Table~\ref{tab:data-stats} provides statistics detailing the quantity of synthetic examples in our dataset which were derived from reasoning failures on different benchmarks. Additional discussion of the dataset composition is provided in Section~\ref{sec:dataset-details}.

\begin{table}[htbp]
    \centering
    \resizebox{\columnwidth}{!}{
    \begin{tabular}{lccc}
        \toprule
        \textbf{$\textrm{Dataset}_{\textrm{Image Type}}$} & \textbf{Original} & \textbf{Failures} & \textbf{Filtered}\\
        \midrule
        $\textrm{VizWiz}_{\textrm{real}}$ & 20,523 & 7,785 & 100,280\\[2pt]
        $\textrm{VizWiz}_{\textrm{syn}}$ & 20,523 & 7,785 & 190,172\\[2pt]
        $\textrm{InfoVQA}_{\textrm{real}}$     & 10,074 & 5,250 & 95,783\\[2pt]
        $\textrm{ScienceQA}_{\textrm{real}}$   & 5,585  & 1,562 & 39,090\\[2pt]
        $\textrm{OK-VQA}_{\textrm{syn}}$ & 9,009  & 607   & 128,667\\
        \bottomrule
    \end{tabular}}
    \caption{Dataset statistics across benchmarks, including original training set size, number of failure samples (LLaVA-1.5-7b: 0, GPT-4o: 1), and synthetic samples with filtering score 3.}
    \label{tab:data-stats}
\end{table}

\subsection{Data generation prompt}
\label{app:generation-prompt}
Figure \ref{prompt:prompt_method1} shows the prompt used to generate fully synthetic question-answer, and image samples based on the failure modes of an LMM, according to Method 2.

\begin{figure}[th]
\begin{tcolorbox}
    \footnotesize
\texttt{You are analyzing the performance of a vision-language model (called Model A). Model A’s answer could deviate from the ground truth due to limitations in visual understanding, interpretation, or reasoning.}

\texttt{\textbf{Step 1}: Describe the image.}

\texttt{\textbf{Step 2}: Given a question, the Ground truth answer, and Model A's generated answer, describe any key visual elements that might influence Model A's interpretation.}

\texttt{\textbf{Step 3}: Analyze the reasoning steps Model A might have used to generate its answer, considering both the visual and textual information. Identify any weaknesses, errors, or gaps in Model A response compared to the ground truth.}

\texttt{\textbf{Step 4}: Suggest 10 additional challenging detailed examples to address these limitations.}

\texttt{\textbf{Step 5}: Transform each example into a detailed prompt designed to generate a clear and realistic image using a text-to-image generation model.}
\end{tcolorbox}
\caption{Prompt used to generate fully synthetic image-text samples based on the failure modes of an LMM (Method 2).}
\label{prompt:prompt_method1}
 \end{figure}

\subsection{Filtering prompt}
\label{app:filtering-prompt}

Figure~\ref{fig:filtering_prompt} provides the prompt which we used for the filtering stage of our synthetic data generation pipeline. See Section~\ref{sec:filtering} of the main paper for additional filtering details.

\begin{figure}[th]
\begin{tcolorbox}
    \small
\texttt{Given sample containing an image, a question, and an answer, your task is to grade the sample from 1 to 3 based on the following criteria:}

\texttt{\textbf{Score 1}: The answer is incorrect.}

\texttt{\textbf{Score 2}: The answer is correct, but it is one of several possible valid answers.}

\texttt{\textbf{Score 3}: The answer is correct, specific, and the only valid answer. The image provides all the necessary context for the answer.}

\end{tcolorbox}
\caption{Filtering prompt}
\label{fig:filtering_prompt}
 \end{figure}

\subsection{Training hyperparameters}
\label{sec:hyperparams}
To train our model, we used 8 Nvidia RTX A6000 GPUs using the hyperparameters from Table \ref{table:hyperp}. We employed DeepSpeed ZeRO stage 3 \cite{deepspeed} for distributed training.

\begin{table*}[ht]
\centering
\begin{tabular}{rrl}
    \toprule
    & Batch Size/GPU & $16$ \\
    & Number of GPUs & $8$ \\
    & Gradient Accumulation & $1$ \\
    & Number of epochs & $1$ \\
    & LLaVA Image Size & $576$ \\
    & Optimizer & AdamW \\
    & Learning Rate & $2e-5$ \\
    & BF16 & $True$ \\
    & LR scheduler & $cosine$ \\
    & Vision Tower & openai/clip-vit-large-patch14-336 \\
    & Language Model & lmsys/vicuna-7b-v1.5 \\
    \bottomrule
\end{tabular}
\caption{Hyperparameters used to train models.}
\label{table:hyperp}
\end{table*}

\section{Additional experimental results}
\label{app:additinal-results}

\subsection{Additional evaluation results for OOD datasets}

We used a variety of multimodal reasoning benchmarks for evaluating models' out-of-domain performance. Since InfoVQA and ScienceQA questions rely heavily on reading text contained in the images, we further evaluated models on TextVQA \cite{singh2019towards} and OCR-Bench \cite{fu2024ocrbenchv2improvedbenchmark}. Finally, because our synthetic dataset was designed to enhance the model's reasoning capabilities, we chose MMBench \cite{liu2024mmbenchmultimodalmodelallaround} and MMMU \cite{yue2024mmmumassivemultidisciplinemultimodal} as additional OOD benchmarks. Evaluation results for the OOD benchmarks are provided in Table~\ref{tab:training-data-augmentation-mix-ood-datasets}. While our primary focus is improving the model's performance in-domain (i.e., correcting its reasoning failures), notably we observe that we are able to achieve this without degrading performance on these OOD tasks. This demonstrates the ability of our approach to correct reasoning failures without sacrificing the model's general reasoning capabilities.

\begin{table*}[h!]
    \centering
    {
    \begin{tabular}{ccclccccccccccc}
    \toprule
    Base LLM & $N$ & $N_{syn}$ & Aug Data & T-VQA & OCR-B & VQAv2 & MMB & MMMU \\
    \midrule
    \multirow{1}{*}{Vicuna-7B} & 624,610 & 0 & N/A (baseline) & 47.0 & 31.9 &71.5 & 52.3 & 36.4\\
    \midrule
    \multirow{4}{*}{Vicuna-7B} & \multirow{4}{*}{687,071} & \multirow{4}{*}{62,461} & ALLaVA & \textbf{47.9} & \textbf{34.0} & \textbf{72.6} & 50.2 & \textbf{36.7}\\
   & & & CoSyn-400K & 47.1 & 31.8 & 72.3 & 52.4 & 34.6\\
  &  & & SimVQA & 46.8 & 31.6 & 71.8 & \textbf{53.5} & 34.7\\
   & & & Ours & 47.4 & 33.2 & 71.3 & 52.5 & 36.2\\
    \midrule
   \multirow{3}{*}{Vicuna-7B} & \multirow{3}{*}{749,532} & \multirow{3}{*}{124,922} & ALLaVA & 47.2 & 34.1 & \textbf{72.8} & 43.5 & 34.2\\
  &  & & CoSyn-400K & 46.8 & 32.7 & 72.0 & 51.9 & 36.9\\
   & & & Ours & \textbf{47.4} & \textbf{34.5} & 72.4 & \textbf{52.5} &  \textbf{37.4}\\
    \midrule
    \midrule
        \multirow{2}{*}{Gemma-2B} & 624,610  & 0  & \multirow{2}{*}{Ours} & 39.9 & 28.3 & 68.7 & 29.2 & \textbf{32.3} \\
     &  749,532 & 124,922 &  & \textbf{40.9} & \textbf{29.9} & \textbf{69.0} &
     \textbf{30.7}  & 31.7\\
    \midrule
   \multirow{2}{*}{Qwen2-7B} & 624,610 & 0 & \multirow{2}{*}{Ours} & 45 & 31.4 &  71.7 &  63 & \textbf{42.3}\\
         &  749,532 & 124,922 &  &\textbf{46.2} & \textbf{32.5} & \textbf{72.9} & \textbf{63.5} & 42.2 \\

    \bottomrule

    \end{tabular}
    }
    \caption{\textbf{Out-of-domain training data augmentation evaluation results.} $N$ denotes the total number of training examples, $N_{syn}$ denotes the number of synthetic examples in the training dataset generated using our approach. T-VQA denotes TextVQA, OCR-B represents OCR-Bench, and MMB means MMBench datasets. The first section of the table compared the same base LLM (Vicuna) trained on various datasets with our dataset, while the second section compares different LLM backbones trained on our dataset.} 
    \label{tab:training-data-augmentation-mix-ood-datasets}
\end{table*}

\subsection{Detailed OOD results for models fit to different subsets of synthetically generated data}

\begin{table*}[]
    \centering
    {
    \begin{tabular}{lcccccccccccc}
    \toprule
    Train Dataset & $N$ & $N_{syn}$ & T-VQA & OCR-B & I-VQA & VQAv2 & OK-VQA & SQA & MMB & MMMU \\
    \midrule
    Baseline & 624,610 & 0 & 0.47 & 0.32 & 0.27 & 0.71 & 0.57 & 0.71 & 52.30 & 0.36  \\
    \midrule
    \multirow{3}{*}{Vizwiz} & 645,133 & 0 & 0.47 & 0.28 & 0.26 &  \textbf{0.72} & \textbf{0.59} & \textbf{0.71} & 51.74 & \textbf{0.38} \\
    & 687,071 & 62,461 & \textbf{0.48} & \textbf{0.32} & 0.26 &  \textbf{0.72} & \textbf{0.59} & \textbf{0.71} & 52.25 & 0.36 \\
    & 749,532 & 124,922 & 0.47 & \textbf{0.32} & \textbf{0.27} & 0.71 & \textbf{0.59} & 0.70 & \textbf{53.02} & 0.37 \\
    \midrule
    \multirow{4}{*}{InfoVQA} & 634,684 & 0 & 0.47 & 0.32 & 0.32 & \textbf{0.73} & 0.58 & 0.70 & 52.50 & 0.37 \\
    & 634,684 & 10,074 & 0.47 & 0.33 & 0.31 & \textbf{0.73} & \textbf{0.59} & 0.70 & 52.16 & 0.36 \\
    & 687,071 & 62,461 & 0.47 & \textbf{0.34} & 0.33 & 0.71 & 0.57 & \textbf{0.71} & \textbf{52.69} & 0.37 \\
    & 710,610 & 86,000 & \textbf{0.48} & 0.33 & \textbf{0.34} &  0.72 & 0.56 & \textbf{0.71} & 52.53 & \textbf{0.38} \\
    \midrule
    \multirow{3}{*}{ScienceQA} & 630,195 & 0 & 0.47 & 0.29 & 0.26 & \textbf{0.73} & \textbf{0.58} & 0.70 & 52.88 & 0.36 \\
    & 630,195 & 5,585 & 0.47 & \textbf{0.32} & \textbf{0.27} & 0.72 & 0.57 & 0.72 & \textbf{53.19} & 0.37 \\
    & 646,594 & 21,984 & 0.47 & \textbf{0.32} & 0.26 & 0.72 & 0.56 & \textbf{0.73} & 53.12 & \textbf{0.38} \\
    \midrule
    \multirow{3}{*}{OK-VQA} & 633,619 & 0 & \textbf{0.47} & 0.30 & 0.27 &  \textbf{0.72} & 0.54 & \textbf{0.71} & \textbf{53.35} &  \textbf{0.36} \\
    & 633,619 & 9,009 & \textbf{0.47} & \textbf{0.33} & \textbf{0.28} & 0.71 & \textbf{0.61} & \textbf{0.71} & 52.68 & 0.35 \\
    & 687,071 & 62,461 & \textbf{0.47} & \textbf{0.33} & 0.27 & \textbf{0.72} & \textbf{0.61} & \textbf{0.71} & 51.96 & 0.35 \\
    \bottomrule
    \end{tabular}
    }
    \caption{\textbf{Training data augmentation experimental results.} $N$ denotes the total number of examples used for training, while $N_{syn}$ denotes the number of synthetic examples in the training dataset which were generated using our approach. T-VQA denotes TextVQA, OCR-B represents OCR-Bench, I-VQA denotes InfoVQA, SQA represents ScienceQA and MMB means MMBench datasets.} 
    \label{tab:training-data-augmentation-ood-full-results}
\end{table*}

Table~\ref{tab:training-data-augmentation-ood-full-results} provides additional evaluation results for models trained individually on real and synthetic data derived from Vizwiz, InfoVQA, ScienceQA, and OK-VQA. All reported values are the official evaluation metrics corresponding to each dataset. The first two rows of each section in Table~\ref{tab:training-data-augmentation-ood-full-results} provide a direct comparison of the efficiency of our synthetic data to real data; we observe that augmenting the LLaVA-Instruct dataset with our synthetic data achieves as good or better performance across most settings as augmenting with real domain-specific data. Furthermore, significant performance gains are achieved relative to the LLaVA baseline when our synthetic data is derived from a dataset in the same domain as the benchmark. For example, synthetic data generated from reasoning failures on InfoVQA significantly improve LLaVA's performance on tasks which require fine-grained text understanding such as OCR-Bench and InfoVQA.

\subsection{Dataset Generation Cost Comparison}
\label{app:cost-comparison}
For 100K synthetic samples, our pipeline required approximately 72 million GPT-4o tokens for question-answer generation and filtering, along with 418 accelerator-hours on Intel Gaudi~2 (or 766 hours on an NVIDIA A100). At current public cloud pricing, this corresponds to roughly US\$0.5K for language model API calls and US\$5.5K for Gaudi~2 compute, totaling approximately US\$6K.

In comparison, CoSyn’s code-guided renderer is dominated by text-only LLM calls, as reported in their paper, generating 100K samples costs approximately US\$2K. The relatively low computational cost of generating chart- and OCR-style images makes CoSyn a natural complement to our method: CoSyn provides structured, code-generated visuals, while our pipeline contributes rich, multimodal failure-supervised content.
ALLaVA uses a slightly simpler caption + question + answer format than ours, omitting our failure-mode identification step. However it provides GPT-4V with high-resolution real images, whereas we downsample to $125 \times 125$ pixels to reduce input token count. Additionally, ALLaVA exclusively uses real images, while we optionally generate synthetic images, adding compute cost in exchange for controllability and diversity.
SimVQA’s Hypersim/TDW pipeline incurs significantly higher generation cost due to photorealistic 3D rendering on GPU clusters. This includes simulating realistic and dynamic environments, making it highly compute-intensive and resource-heavy.
Finally Img-Diff, which generates pairs of images that are visually alike but differ in specific objects. They first generate similar images where some objects are replaced using an LMM and a diffusion model (SDXL \cite{podell2023sdxlimprovinglatentdiffusion} and Prompt-to-prompt \cite{hertz2022prompttopromptimageeditingcross}). Then, they identify the locations of object differences. They extract bounding box regions that contain object differences using CLIP \cite{radford2021learningtransferablevisualmodels} to extract the degree of similiraty, then they use FastSAM \cite{zhao2023fastsegment} to perform image segmentation on each image. Based on the segmentation, they crop the images and use BLIP \cite{li2022blipbootstrappinglanguageimagepretraining} to determine whether an image contains valid objects. Lastly, they use a MLLM to generate descriptive text for the areas with object differences and creates question-answer pairs. In order to run their pipeline and obtain 118k high-quality image pairs, it took four NVIDIA A100 GPUs for 4.5 days.

Overall, our dataset offers a balanced compromise between cost, quality, and diversity, and can be effectively combined with lower-cost image generation pipelines such as CoSyn to enable scalable and semantically rich supervision

\subsection{Generalization to other models}
To evaluate the generalization ability of our synthetic dataset, we trained models using the same datasets but with different backbone LLMs. In the following experiments, we used Gemma-2B \cite{gemmateam2024gemmaopenmodelsbased} and Qwen2-7B \cite{yang2024qwen2technicalreport} as base LLMs.
We adopted the LLaVA two-phase training procedure: pretraining on the LLaVA 558k dataset followed by instruction fine-tuning. Table \ref{tab:training-data-augmentation-mix-all-other-datasets} indicates that our synthetic data in its maximum augmentation setting, $N_{syn} = 124,922$, outperform the baseline for both LLaVA-Gemma-2B and LLaVA-Qwen-7B almost on all benchmarks.

\begin{table*}[]
    \centering
    {
    \begin{tabular}{cclcccccccccccc}
    \toprule
    Model & $N$ & $N_{syn}$  & T-VQA & OCR-B & I-VQA & VQAv2 & OK-VQA & SQA & MMB & MMMU \\
    \midrule
    \multirow{3}{*}{LLaVA-Gemma-2B} & \multirow{3}{*}{624,610} & 0 & 0.3992 & 0.283 & 0.2181 & 0.6866 & 0.5172 & 0.6227 & \underline{29.1752} & \textbf{0.3233} \\
    & & 62,461 & \underline{0.4039} & \underline{0.297} & \underline{0.292} & \textbf{0.6992} & \textbf{0.5536} & \underline{0.6484} & 28.8542  & 0.3144\\
     & & 124,922 & \textbf{0.4092} & \textbf{0.299} & \textbf{0.2981} & \underline{0.6902} & \underline{0.5477} & \textbf{0.6517} &\textbf{30.7322}  & \underline{0.3167}\\
    \midrule
    \multirow{3}{*}{LLaVA-Qwen2-7B} & \multirow{3}{*}{624,610} & 0 & 0.4504 & 0.314 & 0.2674 & 0.7168 & 0.592 & 0.7793 & \underline{62.9986} & \textbf{0.4233}\\
    & & 62,461 & \underline{0.4586} & \underline{0.324} & \textbf{0.2703} & \underline{0.7208} & \textbf{0.6136} & \underline{0.7956} & 62.9900  & 0.4189\\
         & & 124,922 & \textbf{0.4617} & \textbf{0.325} & \underline{0.2699} & \textbf{0.7292} & \underline{0.6064} & \textbf{0.8074} & \textbf{63.4712} & \underline{0.4222} \\
    \bottomrule
    \end{tabular}
    }
    \caption{Training data augmentation experimental results. $N$ denotes the total number of examples used for training, while $N_{syn}$ denotes the number of synthetic examples in the training dataset which were generated using our approach. T-VQA denotes TextVQA, OCR-B represents OCR-Bench, I-VQA denotes InfoVQA, SQA represents ScienceQA and MMB means MMBench datasets. Best result is highlighted in bold and second best result is underlined} 
    \label{tab:training-data-models-generalization}
\end{table*}

\section{Additional analyses}

\subsection{Detailed example comparing different LLMs used for synthetic data generation}
\label{app:llm-comparison}

\begin{figure*}[h] 
    \centering
    \includegraphics[width=\textwidth]{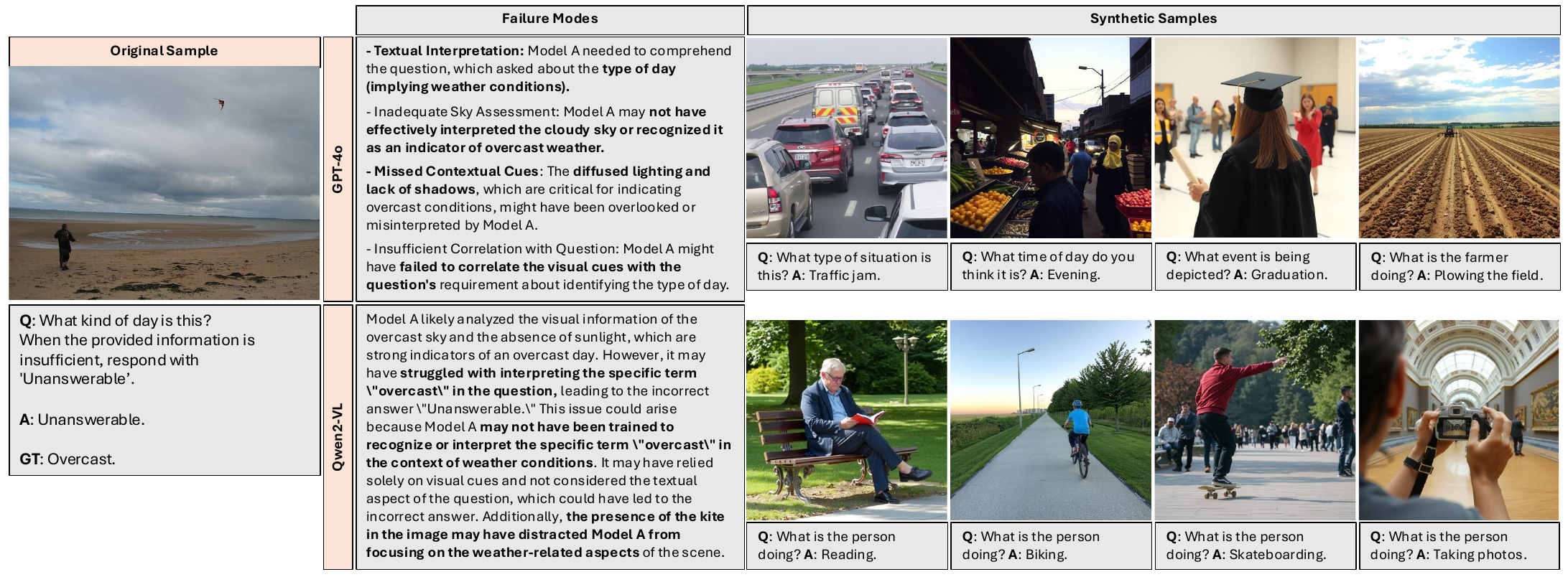}
    \caption{Comparison of GPT-4o and Qwen2-VL for generating Failure-Grounded Synthetic Datasets: GPT-4o demonstrates stronger reasoning capabilities, identifying multiple reasoning failures such as missed contextual cues and a lack of correlation between visual elements and the question. Qwen2-VL also correctly answering the original question, but identifies fewer failure modes. Note: GPT-4o’s reasoning is 2–3 times longer than Qwen2-VL’s; only a portion of GPT-4o’s reasoning is shown here, while Qwen2-VL’s reasoning is presented in full.}
    \label{fig:gpt_qwen_comparison}
\end{figure*}

Figure~\ref{fig:gpt_qwen_comparison} provides a detailed example comparing the reasoning and resulting synthetic examples produced by GPT-4o and Qwen2-VL. See Section~\ref{sec:llm-comparison} for additional discussion.

\section{Examples from our dataset}
\label{sec:illustrative_examples}
\begin{figure*}[h] 
    \centering 
    
    \includegraphics[width=\linewidth]{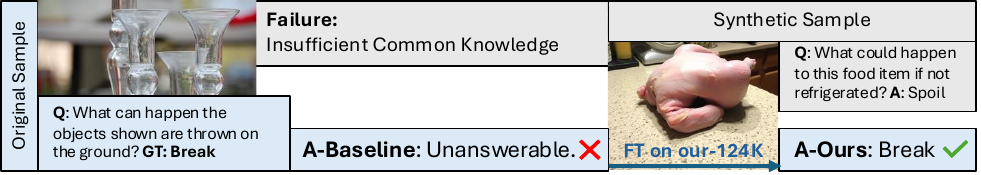}

    \centering 
    \vspace{\intextsep}    \includegraphics[width=\linewidth]{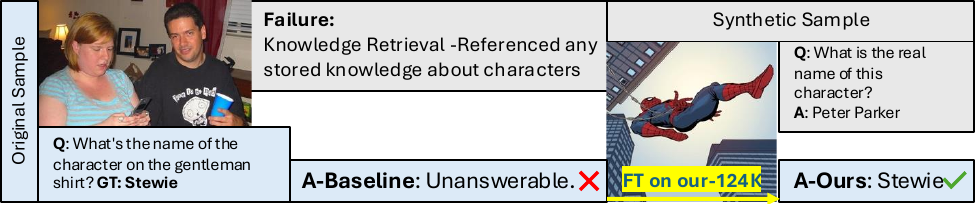}
        \caption{OK-VQA examples with original image-question-answer pairs, the reasoning-based failure identified by GPT-4o, and newly generated counterparts.}
    \label{fig:before_after_example}
\end{figure*}

In this section, we present examples from our dataset and highlight its weaknesses and limitations.
Figure \ref{fig:before_after_example} shows illustrative examples from the OK-VQA dataset. For both samples, we show the original image, question, and answer, followed by the failure mode identified by GPT-4o. We also include newly generated questions, images, and ground truth answers designed to reflect the same failure mechanism, as well as the answer predicted by our finetuned model (on +124k samples).

\begin{figure}[h] 
    \centering
     \includegraphics[width=\columnwidth]{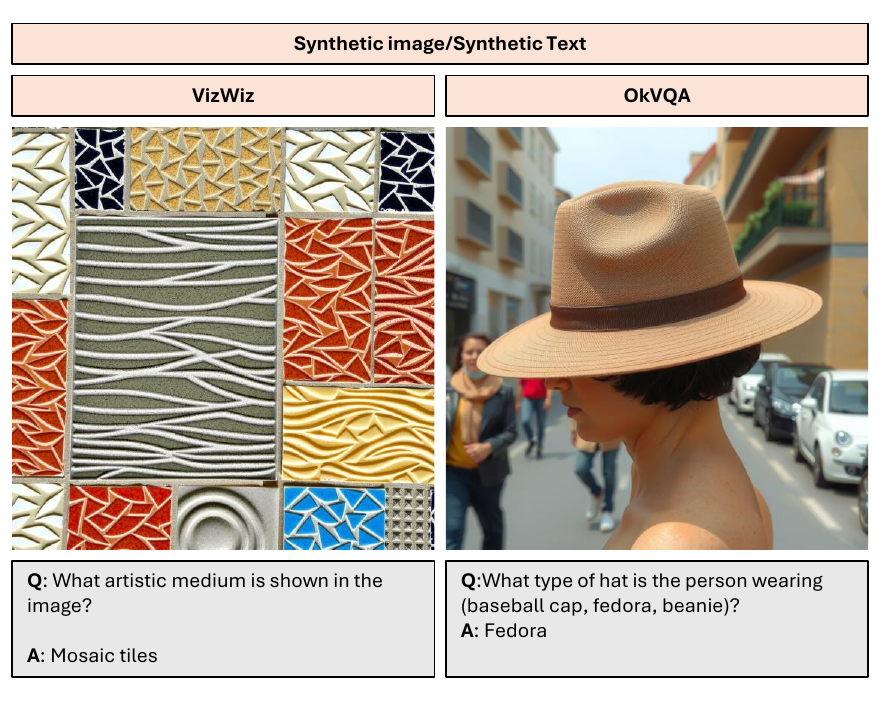}
    \caption{Examples of fully synthetic samples, using Method 2 as described in \ref{sec:synthetic-data-gen}, both question-answer pairs and images were generated.}
    \label{fig:examples_dataset_syn}
\end{figure}

\begin{figure}[h] 
    \centering
     \includegraphics[width=\columnwidth]{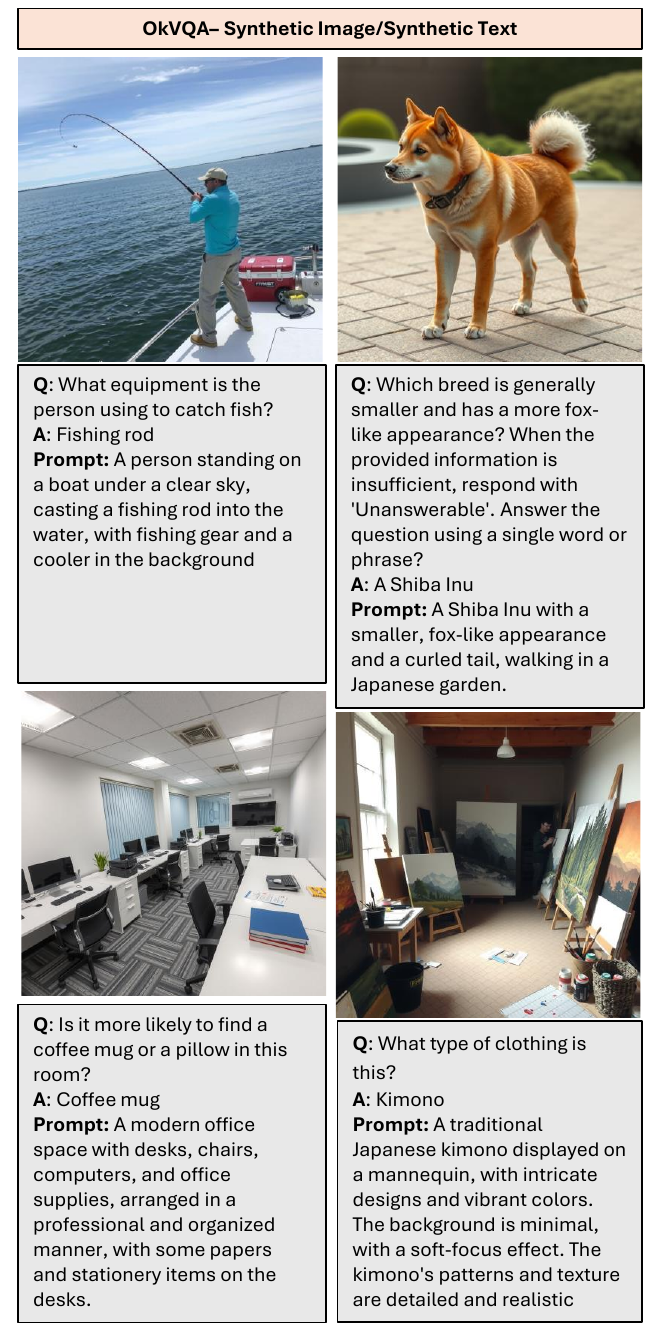}
    \caption{Examples of generated samples from OK-VQA with synthetic images and synthetic text.}
    \label{fig:good_okvqa}
\end{figure}

\begin{figure}[h] 
    \centering
     \includegraphics[width=\columnwidth]{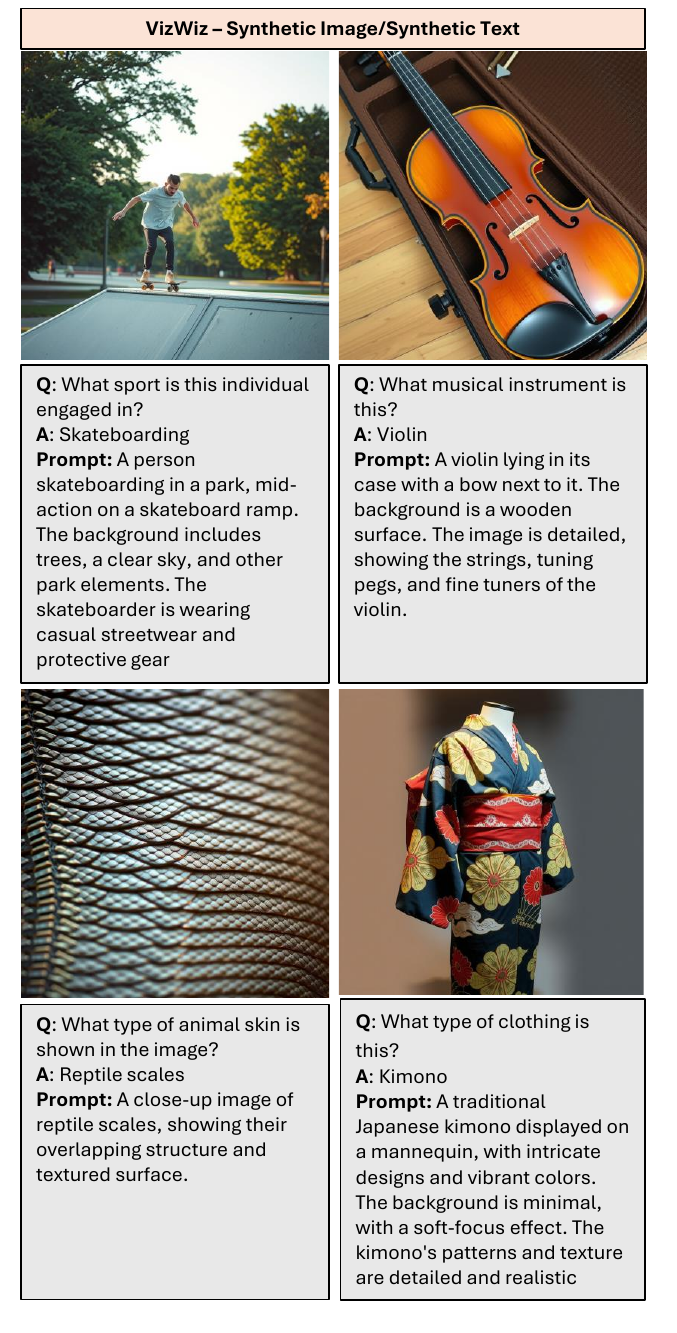}
    \caption{Examples of generated samples from VizWiz with synthetic images and synthetic text.}
    \label{fig:good_vizwizsyn}
\end{figure}

\begin{figure}[h] 
    \centering
     \includegraphics[width=\columnwidth]{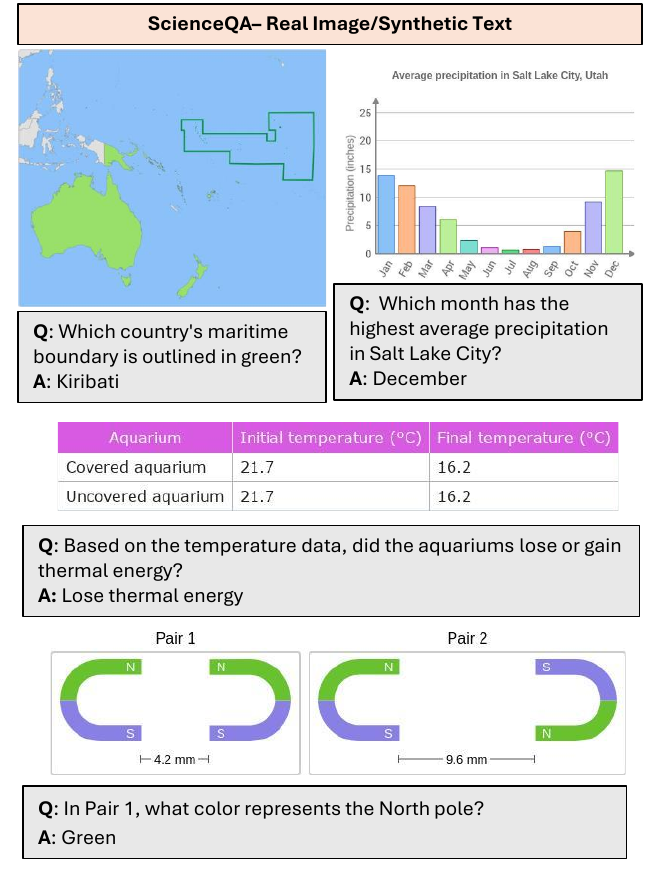}
    \caption{Examples of generated samples from ScienceQA with real images and synthetic text.}
    \label{fig:good_scienceqa}
\end{figure}

\begin{figure}[h] 
    \centering
     \includegraphics[width=\columnwidth]{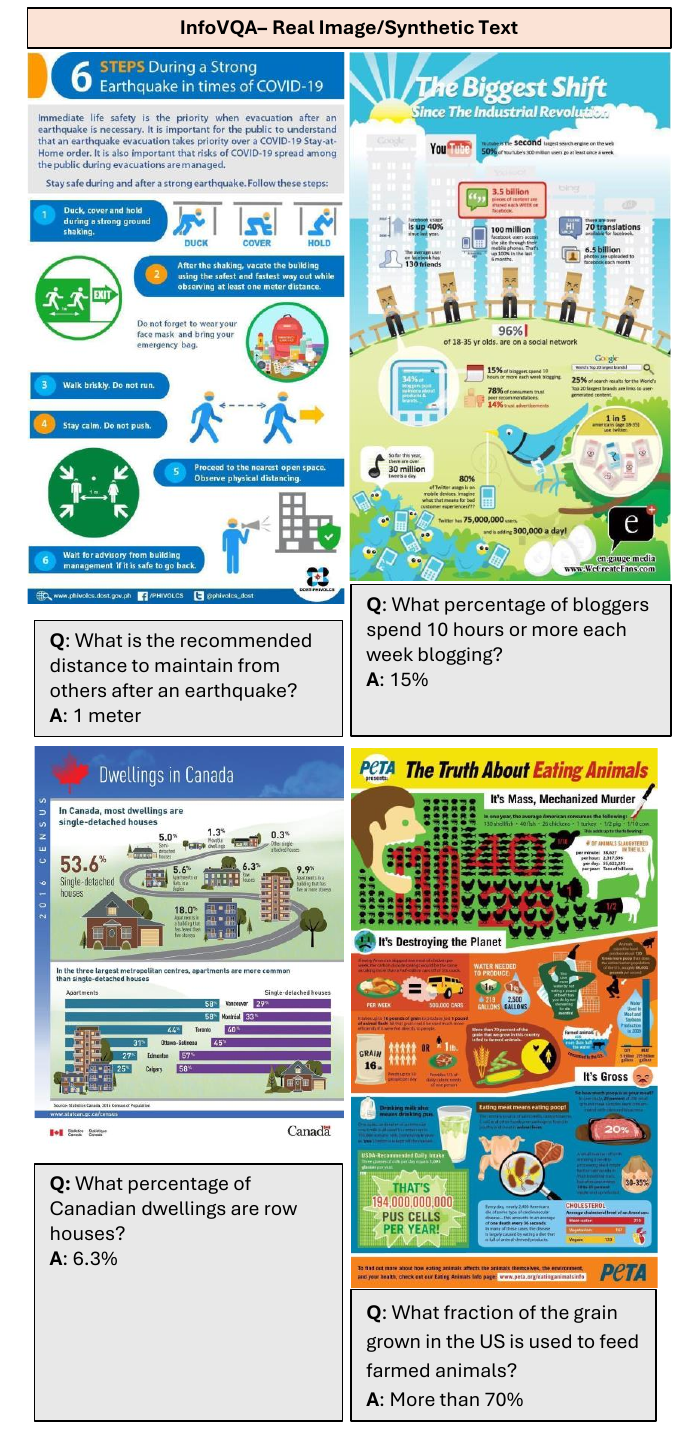}
    \caption{Examples of generated samples from InfoVQA with real images and synthetic text.}
    \label{fig:good_infovqa}
\end{figure}

\begin{figure}[h] 
    \centering
     \includegraphics[width=\columnwidth]{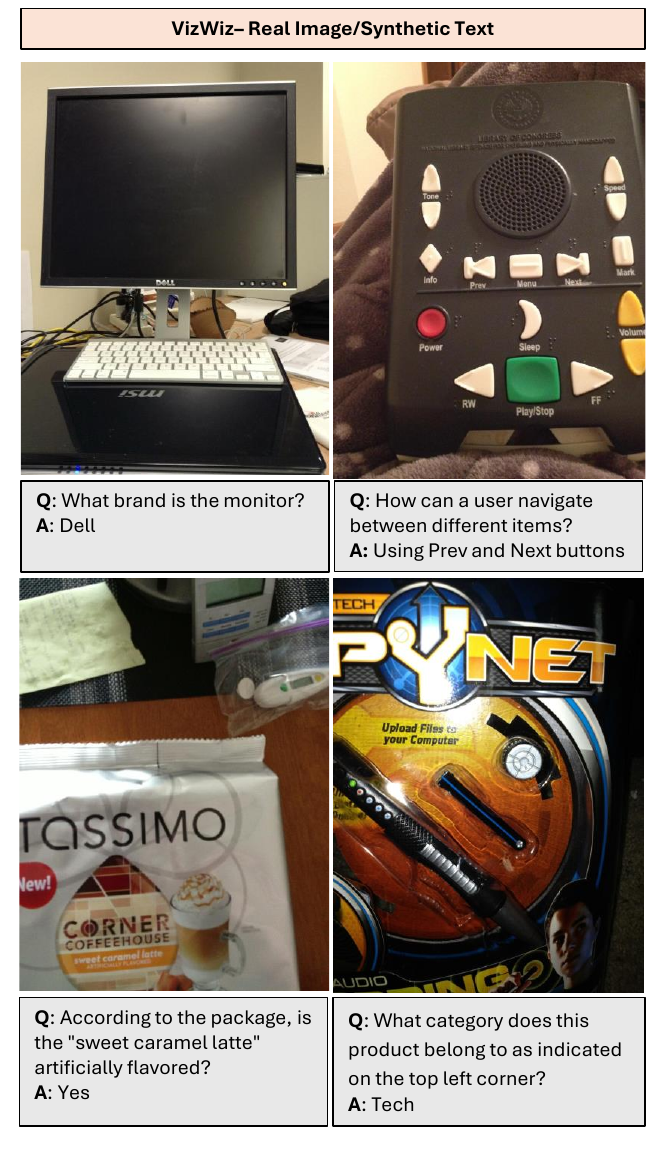}
    \caption{Examples of generated samples from VizWiz with real images and synthetic text.}
    \label{fig:good_vizwizreal}
\end{figure}

\begin{figure}[h] 
    \centering
     \includegraphics[width=\columnwidth]{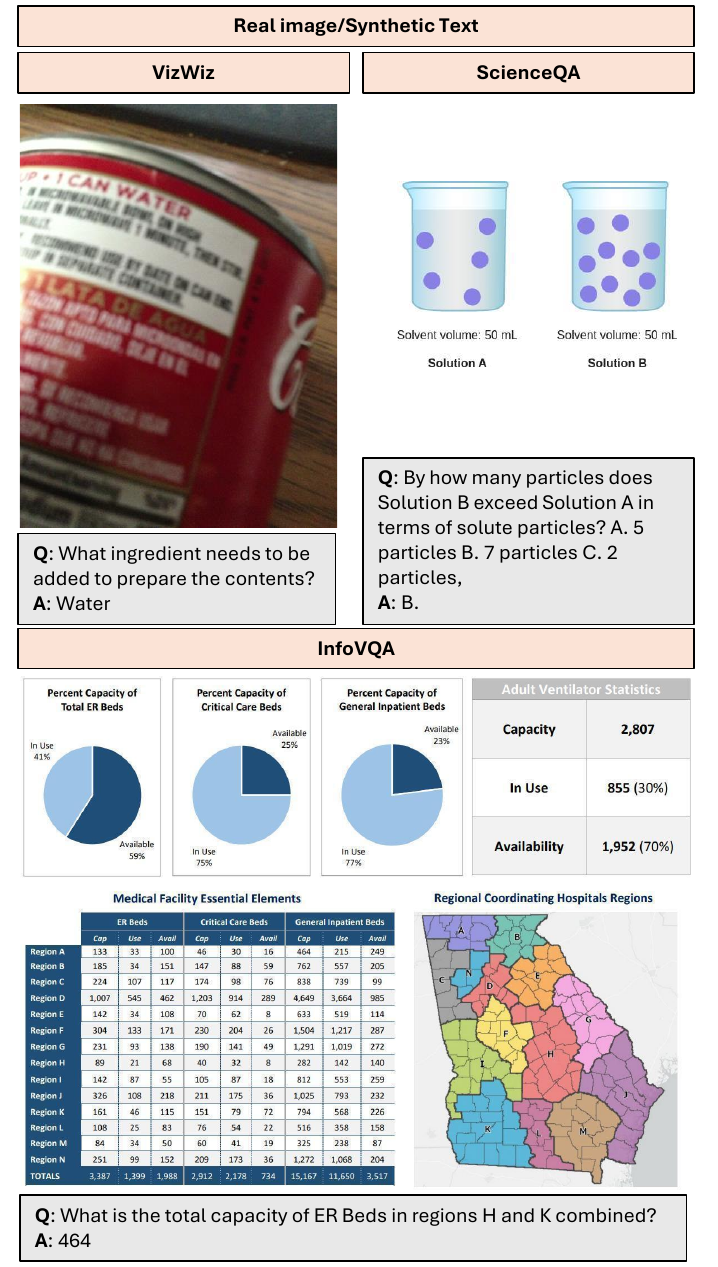}
    \caption{Examples from our dataset real image-synthetic text, where the sample is ambiguous or incorrect.}
    \label{fig:badreal}
\end{figure}

\begin{figure}[h] 
    \centering
     \includegraphics[width=\columnwidth]{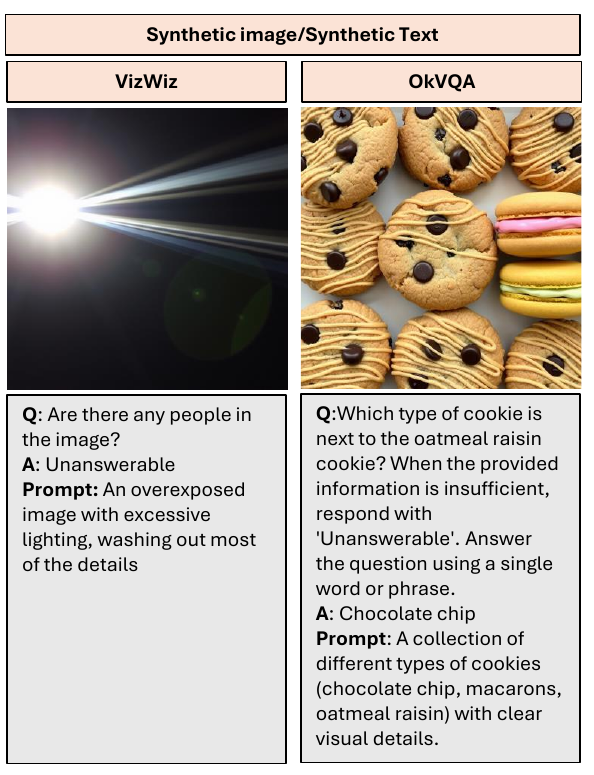}
    \caption{Examples from our dataset synthetic image-synthetic text, where the sample is ambiguous or incorrect. For readability, the ScienceQA image was cropped to focus on the region of interest.}
    \label{fig:badsyn}
\end{figure}

Figure \ref{fig:good_okvqa} shows our synthetic data generated from the OK-VQA dataset while Figure \ref{fig:good_vizwizsyn} corresponds to the VizWiz dataset. In these examples, both the image, the question and the answer are all synthetic. Figure~\ref{fig:examples_dataset_syn} provides additional fully synthetic text \& image examples derived from VizWiz and OK-VQA. Figure  \ref{fig:good_scienceqa}, \ref{fig:good_infovqa} and \ref{fig:good_vizwizreal} illustrate examples derived from the ScienceQA, InfoVQA and VizWiz benchmarks respectively, where the images are real but the questions and answers are synthetically generated. Lastly, Figures \ref{fig:badreal} and \ref{fig:badsyn} show some incorrect examples for each benchmark. 

\subsection{Impact of filtering on data quality}

To investigate the impact of filtering on the quality of synthetically generated data, we repeated our in-domain evaluation experiments for ScienceQA and OK-VQA using raw unfiltered data. In the maximum synthetic data augmentation setting (last row of each section in Table~\ref{tab:training-data-augmentation-in-domain}), using unfiltered data reduces EM from 73.0 to 72.2 on ScienceQA and from 63.3 to 58.8 on OK-VQA. This shows that our filtering approach improves model performance when using our synthetic examples for training data augmentation. Furthermore, using only synthetic examples which were assigned the lowest rating in our filtering process decreases the EM score on OK-VQA to 57.5, which highlights the difference in quality between the lowest-scoring and highest-scoring synthetic examples identified during filtering. 


\subsection{Training data substitution vs. augmentation and impact of synthetically generated images}
\label{app:data-substitution}

\begin{table*}[]
    \centering
    {
    \begin{tabular}{p{2cm}cccccccccccc}
    \toprule
    Train Data & $N$ & $N_{syn}$ & T-VQA & OCR-B & I-VQA & VQAv2 & OK-VQA & SQA & MMB & MMMU \\
    \midrule
    Baseline & 624,610 & 0 & 47.0 & 31.9 & 26.7 & 71.5 & 57.0 & 70.7 & 52.3 & 36.4 \\
    \midrule
    \multirow{2}{2cm}{Substitute w/ syn images} & 624,610 & 62,461 & 47.1 & 31.6 & 26.5 & \textbf{72.9} & 56.9 & 70.8 & 52.3 & 35.3\\
    & 624,610 & 156153 & 46.9 & 31.1 & 27.0 & 72.7 & 57.0 & 70.6 & 51.2 & \textbf{37.9}\\
    \midrule
    \multirow{2}{2cm}{Augment w/ syn images} & 645,222 & 20,612 & 46.7 & 32.0 & 27.0 & 72.6 & 57.4 & \textbf{71.2} & \textbf{53.4} & 34.9\\
    & 687,071 & 62,461 & \textbf{47.7} & 31.8 & 25.8 & 71.6 & \textbf{59.4} & \textbf{71.2} & 52.3 & 36.4\\
    \midrule
    \multirow{2}{2cm}{Augment w/ real images} & 645,222 & 20,612 & 46.9 & \textbf{32.5} & \textbf{27.2} & 72.6 & 57.4 & 70.6 & 53.1 & 35.0\\
    & 687,071 & 62,461 & 47.2 & 32.2 & \textbf{27.2} & 71.8 & 56.9 & \textbf{71.2} & 52.3 & 33.8\\
    \bottomrule
    \end{tabular}
    }
    \caption{\textbf{Ablation experiments comparing baseline LLaVA to LLaVA models trained with synthetic data generated from VizWiz failures.} T-VQA denotes TextVQA, OCR-B represents OCR-Bench, I-VQA denotes InfoVQA, SQA represents ScienceQA and MMB means MMBench datasets. We investigate substitution and augmentation strategies for synthetic data, as well as the use of synthetic vs. real images.}
    \label{tab:training-data-substitution-augmentation-ablations}
\end{table*}

Our previous experiments augmented an existing 624k sample training dataset (LLaVA-Instruct) with our synthetic data. In domains where data is scarce, training datasets of this size may not be available. To investigate the utility of our synthetic data in such low-resource settings, we conducted experiments in which we randomly substituted different quantities of examples from the original dataset with our synthetically generated data\footnote{We used synthetic data derived from Vizwiz failures in this setting.}. 
The results of this experiment are provided in rows 2-3 of Table~\ref{tab:training-data-substitution-augmentation-ablations}. Even when up to 25\% of the original dataset is substituted with our synthetic data, we achieve performance that is either as good or better than the baseline LLaVA model across a broad range of downstream tasks. This is despite the fact that the original LLaVA training dataset utilizes real images, whereas our synthetic data used in this experiment contained only synthetically generated images. The fact that our synthetic data achieves similar or better performance than an existing real data source is significant, as prior studies have shown that training on synthetically generated image data is often much less efficient than training on an equivalent amount of real image data \citep{he2022synthetic}.
Table~\ref{tab:training-data-substitution-augmentation-ablations} also shows the impact of using real vs. synthetic images in our pipeline. Specifically, we compare the effectiveness of our synthetic data derived from Vizwiz reasoning failures when paired with real images (from Vizwiz) or synthetically generated images. In the training data augmentation setting, we observe that synthetic images generally achieve similar results as utilizing real images. Synthetic images even surpass the performance of real images in TextVQA, OK-VQA, and MMBench. This demonstrates the high quality of our synthetic images and their potential to serve as replacements for real images in low-resource settings where data is scarce.

\end{document}